\documentclass[journal]{IEEEtran}
\usepackage{graphicx}
\usepackage{subfigure}
\usepackage{color}
\usepackage{changes}
\usepackage{amsmath}
\usepackage{amsfonts,amssymb}
\usepackage{epstopdf}[outdir=./]
\usepackage{caption}
\usepackage{tikz,xcolor,hyperref}
\usepackage{cite}
\usepackage{float}
\usetikzlibrary{svg.path}
\usepackage[numbers,sort&compress]{natbib}

\definecolor{lime}{HTML}{A6CE39}
\DeclareRobustCommand{\orcidicon}{%
    \begin{tikzpicture}
    \draw[lime, fill=lime] (0,0) 
    circle [radius=0.16] 
    node[white] {{\fontfamily{qag}\selectfont \tiny ID}};    \draw[white, fill=white] (-0.0625,0.095) 
    circle [radius=0.007];    \end{tikzpicture}
    \hspace{-2mm}}
\foreach \x in {A, ..., Z}{%
    \expandafter\xdef\csname orcid\x\endcsname{\noexpand\href{https://orcid.org/\csname orcidauthor\x\endcsname}{\noexpand\orcidicon}}
    }

\begin{document}
\title{Multi-graph Spatio-temporal Graph Convolutional Network for Traffic Flow Prediction}


%
%

\author{
Weilong Ding\orcidB{},\thanks{This work was supported by Beijing Municipal Natural Science Foundation (No. 4192020).}\thanks{Weilong Ding and Tianpu Zhang are at the School of Information Science and Technology, North China University of Technology, Beijing 100144, China, and also in Beijing Key Laboratory on Integration and Analysis of Large-scale Stream Data, Beijing 100144, China.} \and Tianpu Zhang\orcidA{}, 
\and Jianwu Wang\thanks{Jianwu Wang is at Department of Information Systems, University of Maryland Baltimore County, USA}, 
\and Zhuofeng Zhao\thanks{Zhuofeng Zhao is at the School of Information Science and Technology, North China University of Technology, Beijing 100144, China, and also in Beijing Key Laboratory on Integration and Analysis of Large-scale Stream Data, Beijing 100144, China.}
}

%
%


%
\maketitle            
%

\begin{abstract}
Inter-city highway transportation is significant for urban life. As one of the key functions in intelligent transportation system (ITS), traffic evaluation always plays significant role nowadays, and daily traffic flow prediction still faces challenges at network-wide toll stations. On the one hand, the data imbalance in practice among various locations deteriorates the performance of prediction. On the other hand, complex correlative spatio-temporal factors cannot be comprehensively employed in long-term duration. In this paper, a prediction method is proposed for daily traffic flow in highway domain through spatio-temporal deep learning. In our method, data normalization strategy is used to deal with data imbalance, due to long-tail distribution of traffic flow at network-wide toll stations. And then, based on graph convolutional network, we construct networks in distinct semantics to capture spatio-temporal features. Beside that, meteorology and calendar features are used by our model in the full connection stage to extra external characteristics of traffic flow.  By extensive experiments and case studies in one Chinese provincial highway, our method shows clear improvement in predictive accuracy than baselines and practical benefits in business.

\end{abstract}
\section{Introduction}

Inter-city highway plays an important role in urban modern life, its capacity has not been explored enough, and traffic congestion is still one serious issue worldwide. Accordingly, intelligent transportation system (ITS)~\cite{ref_article29,ref_article30} is imperative in highway domain for traffic guidance and travel planning. With the rapid technical development of such as Big Data and Internet of Things, various sensory data has been imported and accumulated in ITS. As one of the basic functions of ITS, traffic flow prediction is widely employed for traffic control, resource dispatching and public guidance.
Traffic prediction method can be short-term, mid-term or long-term according to predictive time period. When less than or equal to 30 minutes, it is a short-term prediction; when greater than 30 minutes and less than one day, it is a mid-term prediction; when exceeds one day, it is a long-term prediction. Meanwhile, according to the spatial range during one prediction, the methods can be for either single-location or multi-location. Daily traffic flow prediction at network-wide toll stations, focused in this paper as an example, is such a long-term and multi-location method.

Over last decade, many solutions have been studied extensively in the perspectives of statistics, machine learning, and deep neural network. However, it still faces challenges to predict daily traffic flow in practice due to inherent limitations. 
First, massive data with imbalanced distribution would deteriorate the performance of prediction at given locations. Classic statistical models, such as ARIMA~\cite{ref_article3} and Kalman
filter~\cite{ref_article40}, only work well on limited samples at limited locations, which are hard to hold performance on huge data. To avoid over-fitting, common strategies of data normalization are required by machine learning models~\cite{ref_article41}, but usually regard extreme volumes at “vital few” locations as outliers. It brings large errors at those pivot stations of highway network. 
Second, the spatial dependencies of highway network imply various semantics, while current works have not been sufficiently explored them. Spatial proximity, widely adopted to build graph topology, indicates that downstream traffic instead of upstream one in highway influences more for future traffic at a certain location~\cite{ref_article39}. In fact, two locations adjacent in cartographic euclidean space, may appear opposite patterns in their traffic flows~\cite{ref_article27}. How to learn different spatial features besides static physical topology is required. 
Third, correlative extra factors are hard to be fully considered, which hinders the predictive effects. For example, implicit calendric periodicity may affect traffic on certain days, such as holidays or weekends; external meteorological conditions, such as heavy snow or heavy rain, may bring similar traffic patterns [2] in highway. All those had to be employed comprehensively for better predictive result.

Accordingly, a method MSTGCN (\textbf{\underline{M}}ulti-graph \textbf{\underline{S}}patio-\textbf{\underline{t}}emporal \textbf{\underline{G}}raph \textbf{\underline{C}}ovolution \textbf{\underline{N}}etwork) is proposed and applied it in a practical business system. The contributions of our work can be summarized as follows. (1) We found long-tail distribution of traffic flow at network-wide toll stations, and propose a strategy to deal with data imbalance. The extreme values of traffic flows at vital few stations are elaborated seriously rather than regarded as outliers.(2) Among highway toll stations, three highway graphs are constructed to learn various spatial semantics of traffic flow. (3) Considering meteorological and and calendric characteristics, spatio-temporal multi-graph fusion with extra factors distinctly improves predictive effects. Moreover, our method has been adopted in a practical ITS and shows convincing benefits with extensive experiments and case studies.

The rest of this paper is organized as follows. Section \ref{section2} shows the related work. Section \ref{section3} demonstrates motivation and problem definition. In Section \ref{section4}, the methodology of our work is elaborated. Section \ref{section5} evaluates the performance and effects by extensive experiments and case studies. In Section \ref{section6}, conclusion is summarized and present the future work.

\section{Related work}\label{section2}
We analyse recent studies into three technical perspectives: traffic flow prediction methods, pre-processing for deep learning methods and service in domain systems.

{\bfseries Traffic flow prediction.} Two raw types of methods exist to predict traffic flow. 
One belongs to statistical methods. These works exploit combinatorial optimization among multivariable factors to improve predictive precision. ARIMA (Autoregressive Integrated Moving Average model)~\cite{ref_article3} and Kalman filtering~\cite{ref_article1} are such popular methods. Regarding data as time series, they have the stationary and linear assumption among temporal characteristics. However, real data is always too complex to satisfy those assumptions, and poor performance appears on huge data in practice. SVR (Support Vector Regression)~\cite{ref_article4}, KNN (K-Nearest Neighbor)~\cite{ref_article6}, Bayesian model~\cite{ref_article8,ref_article9} and GBRT (Gradient BoostRegression Tree)~\cite{ref_article7} have also been widely used in domain due to calculative efficiency. However, they perform not well at some given locations, because such models are sensitive to the quality of training data, and require specific dedicated calibration before prediction.  
The other type is deep learning (DL) methods. As a branch of machine learning, deep learning models can achieve more accurate results without complex feature engineering. RNN (recurrent neural networks), CNN (convolutional neural networks) and their integrated networks are widely used for traffic flow prediction. As variants of RNN, the works~\cite{ref_article10,ref_article11,ref_article59,ref_article60} capture dynamic time relationship of traffics. Based on CNN, the works~\cite{ref_article16,ref_article17,ref_article61,ref_article62} predict  traffic flow by dedicated spatial relationship. As a variant of CNN, GCN (graph convolutional network) like ~\cite{ref_article19,ref_article20,ref_article63,ref_article64} is commonly the best choice nowadays. GCN solutions, such as~\cite{ref_article21,ref_article22}, can achieve pretty good accuracy with the spatial features in euclidean space. More works like~\cite{ref_article23,ref_article24}, combine GCN and RNN to obtain better spatio-temporal characteristics of locations, sensors, and time series. However, such methods rely much on self-learning ability of deep learning models, while don't consider imbalanced data distribution in practice. Moreover, spatial relation in real-world has many different perspectives~\cite{ref_article55,ref_article56}, but those works only get one in a specific semantic (e.g., physical topology or self-learning relationship), which would limit predictive results. 
For daily traffic flow prediction, both types of methods to predict traffic flow trend to focus on traffic flow data itself but ignore the external factors influencing traffic. In fact, meteorologic factors like extreme weather status and periodic calendar factor like holidays, can often effect road condition much. To address these issues, a hybrid deep learning model is proposed in our method with the idea of~\cite{ref_article28}. After elaborate data normalization, three graphs from different spatial perspectives are built to fuse characteristics on heterogeneous data. 

{\bfseries Data pre-processing.} To feed enough qualified input to models, data pre-processing is always required. 
On the one hand, data imputation, also known as data cleaning, aims to fill missing values, smooth noise and remove outliers~\cite{ref_article44}. A data cleaning method is proposed in public transportation analyses~\cite{ref_article45}, and guarantees data consistency and legality through time-based clustering and rule-based filtering. For online data pre-processing in highway domain, a method is designed in edge computing environment~\cite{ref_article3} to efficiently ensures records’ validity and continuity. Data imputation technologies still remain challenges in intelligent transportation systems yet~\cite{ref_article46} to improve data quality. 
On the other hand, data has to be treated carefully before model learning to hold steady performance~\cite{ref_article47}, and data normalization is necessary to transform raw data into fixed range. Although not mentioned intentionally in many studies, some strategies have been adopted to reduce the differences of dimensions, such as standardization~\cite{ref_article48}, Z-score~\cite{ref_article49}, Min-Max~\cite{ref_article50}, and parameter regularization~\cite{ref_article51}. It is especially significant for sparse data. For example, in urban rail domain, sparse OD data is in long-tail distribution: in one day, more than 40\% OD pairs is zero; OD flows fewer than two account for more than 65\%. However, extreme values in distribution are always regarded as outliers by traditional normalization strategies, which brings much larger predictive errors at some locations~\cite{ref_article54}. Therefore, normalization strategies have to be re-considered when data is in imbalanced distribution~\cite{ref_article53}. In our solution, besides pre-processing for feature engineering, traffic flow data is normalized elaborately to fit practical long-tail distribution, which improves predictive accuracy at some "vital few" stations in highway.

{\bfseries Prediction applied in transportation domain.} Traffic flow, traffic speed and traffic demand are three basic functions in intelligent transportation system. Specifically, over a certain period of time, traffic flow counts the number of vehicles at a given location; traffic speed represents average driving speed of all the vehicle passing by given road segment; traffic demand expects to uncover the demands for shared transportation at given areas. All those predictive functions have been studied and adopted in ITS nowadays, with loose-coupling service methodology. For simulation, MATSim (multi-agent transport simulation framework)~\cite{ref_article36} service is to model traffic conditions in large-scale urban environment. Based on MATSim, the works ~\cite{ref_article37,ref_article38,ref_article42} support traffic demand analyses for shared transportation, such as ride-pooling, ride-hailing and taxis. Incorporating data processing techniques, MOBDA~\cite{ref_article42} as a microservice-oriented system supports predictive modelling and service analytics for smart cities. CO-STAR~\cite{ref_article6} is a traffic flow prediction service for highway transportation, but it is for 5-minute short-term prediction without considering external daily factors. On heterogeneous daily data including traffic flow, calendar and weather status, our work in this paper is for a one-day long-term traffic flow prediction, and has employed more features to improve performance in a practical system.

\section{Preliminary}\label{section3}

\subsection{Motivation}\label{section3.1}

Our work originates from \textit{Highway Big Data Analysis System} running in Henan, the most populated province in China. The system we built has been in production since October 2017 to improve routine business analytics for highway management. Operated by \textit{Henan Transport Department}, a billion records of heterogeneous data in recent two years have been imported into the system, such as meteorological data, solar / lunar calendric data, real-time license plate recognition data, and toll data. A record of toll data is generated from a device at toll station when a vehicle is charged. As the typical spatio-temporal data, the records of such toll data could be aggregated into traffic flow data in various temporal granularities. The data used in this paper is traffic flows at network-wide toll stations on days.

We revisit daily traffic flows at toll stations in Henan highway, and found interesting observations below. 

\begin{figure*}[htbp]
\begin{center}
\centering
\subfigure[ ]{
\label{fig_flow_status.sub.1}
\begin{minipage}[t]{0.5\linewidth}
\centering
\includegraphics[width=2in]{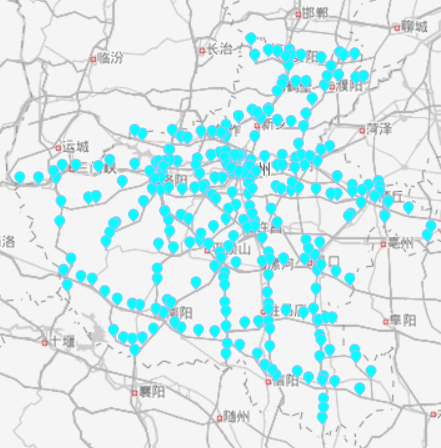}
\end{minipage}%
}%
\subfigure[ ]{
\label{fig_flow_status.sub.2}
\begin{minipage}[t]{0.5\linewidth}
\centering
\includegraphics[width=3.5in]{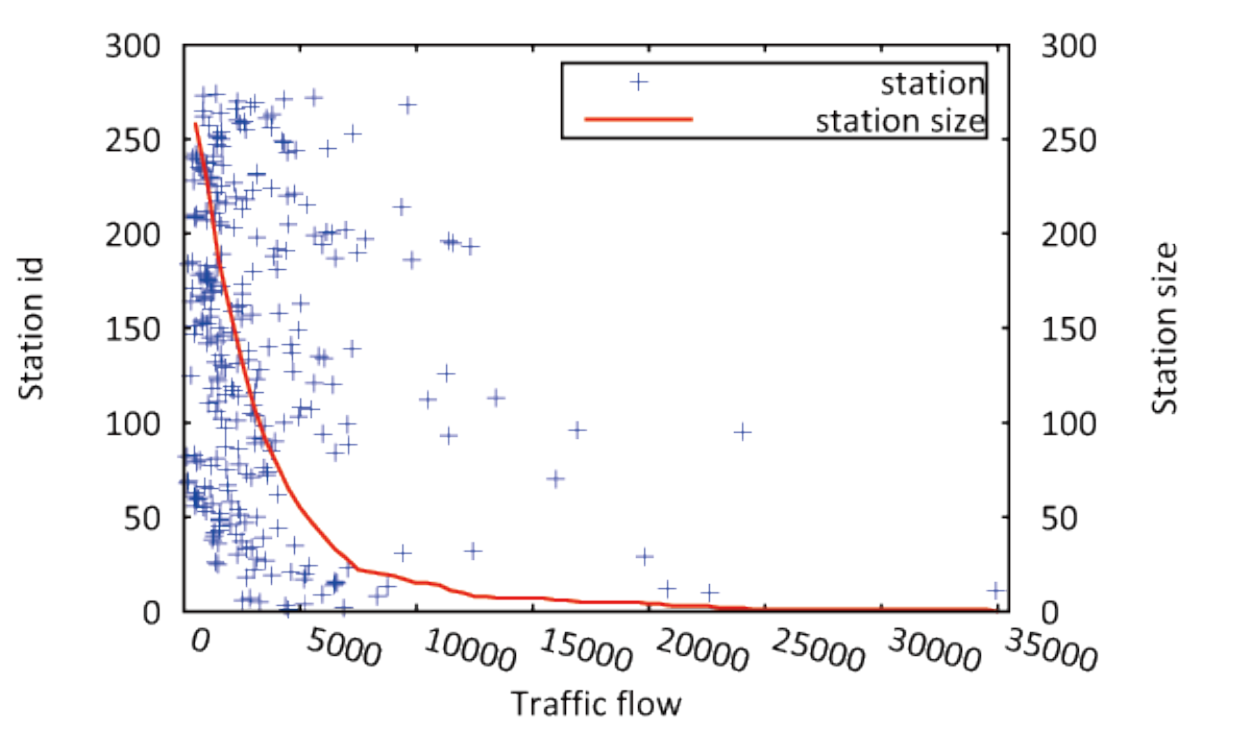}
\end{minipage}%
}%
\centering
\caption{Toll stations and their traffic flow distribution}
\label{fig_flow_status}
\end{center}
\end{figure*}

\textbf{Observation 1}. Among 269 toll stations, the daily traffic flows are extreme imbalanced and appear long-tail distribution (also known as Pareto principle or “80/20” rule). In the system we developed, toll stations are dotted in the provincial map as Figure \ref{fig_flow_status.sub.1}, and their traffic flows on a certain day of year 2018 are presented as Figure \ref{fig_flow_status.sub.2}. In Figure \ref{fig_flow_status.sub.2}, each blue scatter shows the traffic flow at one station; the red line counts stations whose traffic flow larger than the value of x-coordinate. The top toll station \textit{ZhengzhouSouth} has tenfold volume than the average among others in highway network. The top-10 stations occupy nearly 20\% traffic flows, among which six ones are around city \textit{Zhengzhou}, the provincial capital of Henan Province. 

\textbf{Observation 2}. Daily traffic flow is affected by calendar and weather conditions. On weekends or holidays, traffic flow at the stations lying in popular tourism regions would burst due to more private cars’ arriving. Under extreme meteorological condition, such as heavy rain or heavy snow, the traffic in highway would be strictly controlled even closed by officials considering drivers’ safety.

\textbf{Observation 3}. During traffic flow prediction, the toll stations with large traffic flow also own large predictive errors. Current models prone to choose the solution optimally fitting the ensemble, and not properly reflect the facts at specific ones. Therefore, at those busy stations, such as the ones surrounding city \textit{Zhengzhou} with much larger traffic flows, the deviation from other majorities brings larger error in their predictive results.

Accordingly, a novel method is required for traffic flow prediction to improve predictive accuracy. It is just our original motivation.

\subsection{Problem analysis}\label{section3.2}
According to the motivation above, we can abstract the problem as follows. To predict daily traffic flows at network-wide toll stations, our method is to find a function $F(\cdot)$, which can map the graph signals on historical $h$ days to the graph signals on incoming $f$ days. It is described as equation (\ref{equation_problem}).
\begin{equation}
\begin{split}
(y_{d+1},\cdots,y_{d+f})=&F(y_{d},\cdots,y_{d-h+1};G)
\label{equation_problem}
\end{split}
\end{equation}

Here, graph $G=(\mathcal{V}, \mathcal{E}, W)$ depicts the spatial structure of the toll stations in highway network. $\mathcal{V}$ is the set of vertices representing toll stations, $\mathcal{E}$ is the set of edges between any pair of toll station, and $W\in\mathbb{R}^{V*V}$ is the adjacency matrix counting the weights of station pairs. $V=\lvert\mathcal{V}\rvert$ is the number of vertices. The graph signal $y\in{\mathbb{R}^{V*S}}$ is measured on graph $G$, where $S$ is the feature dimension of each vertex on graph $G$. $y_{d}$ represents the value of each vertex on the graph on the day $d$. 

\section{Method}\label{section4}
\subsection{Overview}\label{section4.1}
\begin{figure*}[htbp!]
\begin{center}
  \includegraphics[width=0.75\textwidth]{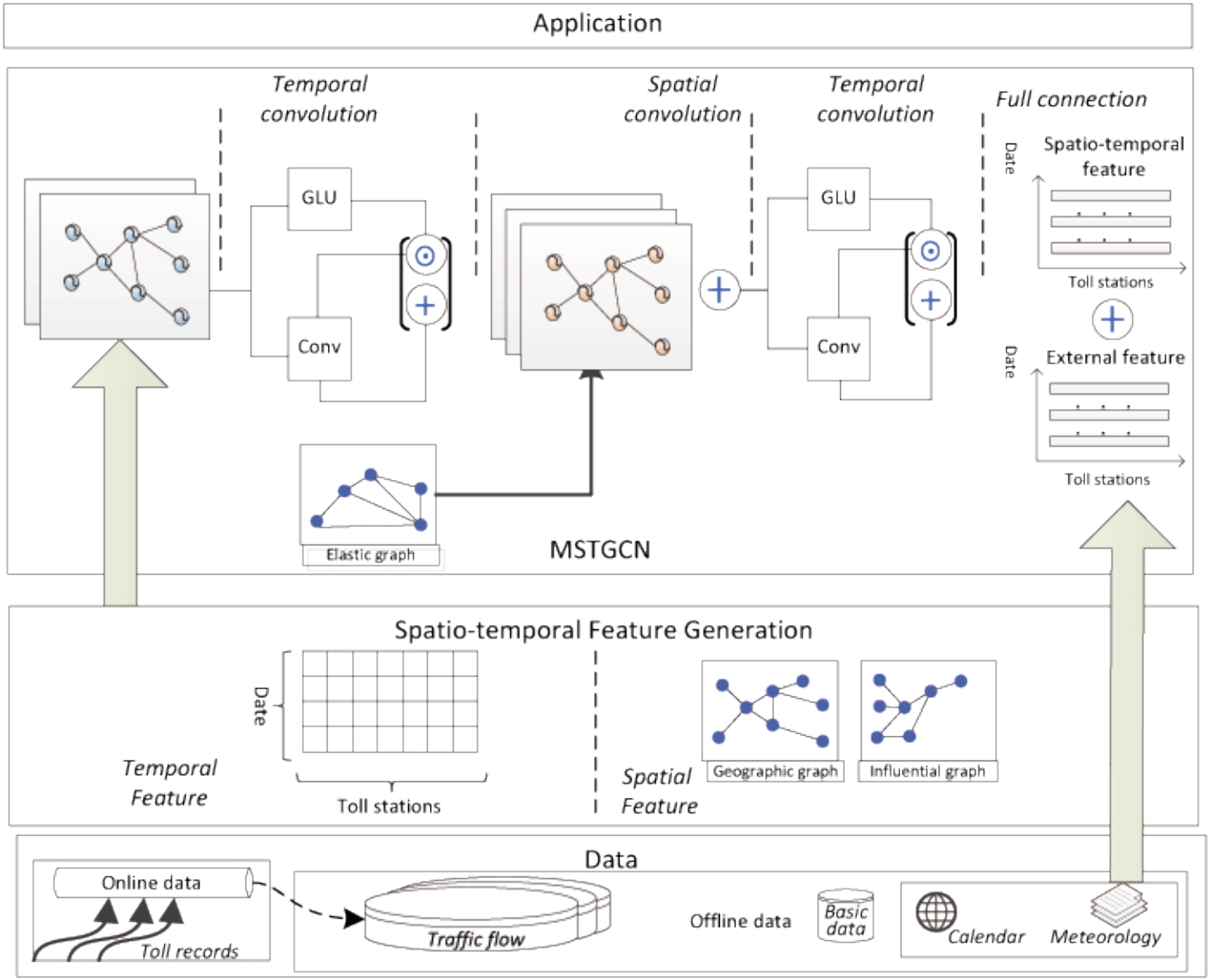}
  \caption{Overview}
  \label{structure}
\end{center}
\end{figure*}

According to the domain requirements in Section \ref{section3.1}, we propose a method MSTGCN (\textbf{\underline{M}}ulti-graph \textbf{\underline{S}}patio-\textbf{\underline{t}}emporal \textbf{\underline{G}}raph \textbf{\underline{C}}onvolution \textbf{\underline{N}}etwork) to predict daily traffic flow. Figure \ref{structure} shows the framework of MSTGCN. 
The input of our method is online and offline data. Raw records of toll data are received continuously through a message broker, and then aggregated as traffic flow data into No-SQL database. Related data cleaning and aggregative calculation can be referred in our previous works~\cite{ref_article6, ref_article57}. From dedicated external data sources, calendar and meteorology data are imported periodically into a relational database. Business basic data, such as profiles of station, section and highway line, has been maintained in that relational database. 

On such heterogeneous data, MSTGCN includes four stages. In the feature processing stage,  external data (i.e., date and weather) is labelled and imbalanced traffic flow data is normalized . All those and would be discussed in Section \ref{section4.2}. In the temporal convolution stage, the temporal series characteristics of traffic flow are extracted by convolution operations, where the forgetting mechanism is adopted through GLU (gated linear units) module~\cite{ref_article31}. In the spatial convolution stage, the spatial characteristics of highway network are captured from different perspectives, and then merged by feature fusion. In the full connection stage, spatial-temporal characteristics and external factors are comprehensively employed through a fully connected network to output predictive results. With the model defined in Section \ref{section4.3}, those stages would be elaborated in Section \ref{section4.4}. 

As a routine method in the system mentioned in Section \ref{section3.1}, the trained model would execute once a day at 12:00 a.m. Some domain applications can employ those results to complete business requirements, such as analytical visualization and potential hot-spots discovery.

\subsection{Data pre-processing}\label{section4.2}
In the feature processing stage of MSTGCN, data pre-processing is done to build external feature and complete data normalization. 

{\bfseries External feature on meteorologic \& calendaric data.} Referred~\cite{ref_article7}, external factor $P_\mathcal{D}^\mathcal{V}=(Q_\mathcal{D}^\mathcal{V}, R_\mathcal{D})$ at network-wide stations $\mathcal{V}$ on days $\mathcal{D}$ is defined as the combination of weather conditions $Q_\mathcal{D}^\mathcal{V}$ and calendric type $R_\mathcal{D}$. Here, $\lvert\mathcal{D}\rvert=D$, $\lvert\mathcal{V}\rvert=V$. On the day $d \in \mathcal{D}$ at a toll station $v \in \mathcal{V}$, the external feature can be expressed as $p_{d}^{v}=(q_{d}^{v},r_{d})$.

As the equation (\ref{equation_weather}), weather condition here is depicted in two categories: one is extreme weather including heavy rain, heavy fog, and strong wind, and the others are normal weather. In extreme weather condition, traffic in highway always would always be restricted.

\begin{equation}
q_{d}^{v}=
\left\{
\begin{array}{rcl}
0,\quad & normal\, weather\,at\,v\,on\,d\\
1,\quad & extreme\,weather\,at\,v\,on\,d
\end{array}
\right.
\label{equation_weather}
\end{equation}

As the equation (\ref{equation_date}), calendric type $R_D$ here are defined by label encoding as three types. During holiday and weekend, traffic would burst at some popular locations. 

\begin{equation}
r_{d}=
\left\{
\begin{array}{rcl}
0,\quad & d \,is \,a\,workday\\
1,\quad & d \,is \,a\,holiday\\
2,\quad & d \,is\, a\, weekend
\end{array}
\right.
\label{equation_date}
\end{equation}

{\bfseries Data normalization on traffic flow data.}  
Data normalization is required to re-scale the range of data. As Observation 1 of Section \ref{section3.1}, long-tail distribution appears on traffic flow data $Y^\mathcal{V}_\mathcal{D}\in \mathbb{R}^{V*D*S}$ at toll stations $\mathcal{V}$ on days $\mathcal{D}$, where $S$ is the dimensionality of features. At some toll stations, the greater values of traffic flow is, the larger predictive errors would be. To emphasize values of vital few rather than discarding them as outliers, Box-Cox transformation is adopted here as normalization strategy. It enables data transformation to meet variance and normality without losing much crucial information. 
As equation (\ref{equation_box_cox}), traffic flows are re-scaled, which reduces the correlation between unobservable errors and predictors to a certain extent. Moreover, to realize end-to-end prediction, the inverse transformation as equation (\ref{equation_box_cox_reverse}) would be used in the last stage of our method . Here, $Y^{\lambda}$ is the output after transformation with the same dimension as the input $Y^\mathcal{V}_\mathcal{D}$. 

\begin{equation}
Y^{(\lambda)}=
\left\{
\begin{array}{rcl}
\frac{Y^\lambda-1}{\lambda},\quad & \lambda \neq 0\\
\ln{(\lambda)},\quad & \lambda = 0
\end{array}
\right.
\label{equation_box_cox}
\end{equation}

\begin{equation}
Y=
\left\{
\begin{array}{rcl}
(1+\lambda Y^{(\lambda)})^{\frac{1}{\lambda}},\quad & \lambda \neq 0\\
\exp(Y^{(\lambda)}),\quad & \lambda = 0
\end{array}
\right.
\label{equation_box_cox_reverse}
\end{equation}

Here, parameter $\lambda$ has to be determined on traffic flows at all toll stations through Box-Cox log-likelihood function. It is computationally cumbersome to test-and-compare when huge data involved. To find the best $\lambda$ efficiently, a trick is adopted here. Only the days when top-K values occur, instead of all $D$ days, are considered to calculate log-likelihood and find the maximum. The positive integer $K$ is not larger than 5 in MSTGCN. Without loss of bound values of traffic flows, our strategy is sound enough for such imbalanced values. 
In fact, vital few stations can be further distinguished from others in long-tail distribution: the toll stations, whose traffic flow is more than three times than that standard deviation on those $K$ days, are regarded as vital few ones according to three-sigma rule. 
%

\subsection{Multi-graph modeling for various spatial semantics}\label{section4.3}

As mentioned in Section \ref{section3.2}, $G=(\mathcal{V}, \mathcal{E}, W)$ is the graph of the highway network. Between $v_i\in\mathcal{V}$ and $v_j \in\mathcal{V}$, $e_{ij}\in{\mathcal{E}}$ is a specific road segment and $w_{ij}\in W$ is directed weight, where $i,j$ are positive integer. 
According to various spatial perspectives, three graphs of highway network are defined here in multiple semantics, including geographic graph ($G_g$), influential graph ($G_r$) and elastic graph ($G_s$).

{\bfseries Geographic graph $G_g$.} 
From physical perspective, among toll stations $\mathcal{V}$ and edges $\mathcal{E}$, geographic graph $G_g=(\mathcal{V}, \mathcal{E},W_g)$ depicts physical semantics of topological connectivity in highway network, and $W_g$ is the weight of $e_{ij}$. When $V=\lvert\mathcal{V}\rvert$, the adjacency matrix $W_g$ of $G_g$ is defined as equation (\ref{equation_Wg}), and its element $W_g(i,j)$ is showed as equation (\ref{equation_Wg_ij}) indicating whether a toll station $i\in\mathcal{V}$ is directly connected to the toll station $j\in\mathcal{V}$. 
\begin{equation}
W_g=
\begin{pmatrix}
    0&W_g(1,2)&\cdots&W_g(1,V)\\
    W_g(2,1)&0&\cdots&W_g(2,V)\\
    \vdots&&\ddots&\vdots \\
    W_g(V,1)&W_g(V,2)&\cdots&0
\end{pmatrix}
\label{equation_Wg}
\end{equation}

\begin{equation}
W_g(i,j)=
\left\{
\begin{array}{rcl}
0,\quad & i,j\,\, is\,\, not\,\, directly\,\, connected\\
1,\quad & i,j\,\, is\,\, directly\,\, connected\\
\end{array}
\right.
\label{equation_Wg_ij}
\end{equation}

{\bfseries Influential graph $G_r$.}
From the statistical perspective in business, among toll stations $\mathcal{V}$ and edges $\mathcal{E}$, influential graph $G_r=(\mathcal{V}, \mathcal{E},W_r)$ presents the influence of a toll station for others in highway network, and $W_r$ represents the weight of $e_{ij}$. 
The influence of any station pair can be counted by historical statistics~\cite{ref_article9}. Based on the analysis in~\cite{ref_article6}, $W_r$ is defined as an asymmetric adjacency matrix in equation (\ref{equation_Wr}), and its element $W_r(i,j)$ as equation (\ref{equation_Wr_ij}) shows the normalized value of influence between toll stations $i,j$. 
In equation (\ref{equation_Wr_ij_scal}), $dis(i,j)$ is the cartographic distance between toll station $i$ and $j$, and $mileage(i)$ is the average mileage of vehicles exiting toll station $i$. Here, $\mathcal{J}$ is toll stations except the focused $i$. 
The influential graph $G_r$ depicts statistical semantics through the upstream dependency of highway network: a vehicle exiting a location must have entered highway in an upstream one, therefore the exit traffic flow (default traffic flow in domain) is influenced by the the most dependent ones. 

\begin{equation}
W_r=
\begin{pmatrix}
    0&W_r(1,2)&\cdots&W_r(1,V)\\
    W_r(2,1)&0&\cdots&W_r(2,V)\\
    \vdots&&\ddots&\vdots \\
    W_r(V,1)&W_r(V,2)&\cdots&0
\end{pmatrix}
\label{equation_Wr}
\end{equation}

\begin{equation}
W_r(i,j)=\frac{\exp{(scal(i,j))}}{\sum_{i \neq j}\exp{(scal(i,j))}}
\label{equation_Wr_ij}
\end{equation}

\begin{equation}
\begin{aligned}
&scal(i,j)=\\
&1-\frac{|dis(i,j)-mileage(i)|-min(|dis(i,\mathcal{J})-mileage(i)|)}{max(|dis(i,\mathcal{J})-mileage(i)|)}
\label{equation_Wr_ij_scal}
\end{aligned}
\end{equation}

{\bfseries Elastic graph $G_s$.}
From the latent relation perspective, among toll stations $\mathcal{V}$ and edges $\mathcal{E}$, elastic graph $G_s=(\mathcal{V}, \mathcal{E},W_s)$ implies inherent spatial relationship by self-learning, and $W_s$ represents the weight of $e_{ij}$. 
The adjacency matrix $W_s$ is defined as equation (\ref{equation_Ws}), would to be re-trained periodically. Its element $W_s(i,j) \in{W_s}$ is the inherent relation between toll stations $i$ and $j$ learned by full-connection neural network. In equation (\ref{equation_Ws_ij}), $(a_1 \ a_2\ \cdots\ a_i \cdots\ a_{V})^\top$ represents the weight vector, and each $a_i \in(0..1)$ for a toll station $i$ is initialized from a normal distribution. Through gradient descent and back propagation, $a_i$ would be gradually updated until the spatial relation is learned.  
Different with $G_g$ and $G_r$ above, whose adjacency matrix are somewhat static, the elastic graph $G_s$ depicts dynamic inherent semantics of highway network through self-learning model without prior knowledge.

\begin{equation}
W_s=
\begin{pmatrix}
    0&W_s(1,2)&\cdots&W_s(1,V)\\
    W_s(2,1)&0&\cdots&W_s(2,V)\\
    \vdots&&\ddots&\vdots \\
    W_s(V,1)&W_s(V,2)&\cdots&0
\end{pmatrix}
\label{equation_Ws}
\end{equation}

\begin{equation}
W_s(i,j)=a_i*a_j  \quad i,j\in \mathcal{V}
\label{equation_Ws_ij}
\end{equation}

{\bfseries Problem definition.} With graphs $G_g$, $G_g$, $G_s$ of highway network from various perspectives and external feature $P_\mathcal{D}^\mathcal{V}$, the traffic flow prediction problem traditionally presented as equation (\ref{equation_problem}) can be transformed to equation (\ref{equation_problem_definition}) in our work.

\begin{equation}
\begin{split}
(\hat{y}_{d+1}^\mathcal{V},\hat{y}_{d+2}^\mathcal{V},\cdots,\hat{y}_{d+f}^\mathcal{V})=&F(y_{d}^\mathcal{V},y_{d-1}^\mathcal{V},\cdots,y_{d-h+1}^\mathcal{V};\\
&P_{d}^\mathcal{V},P_{d-1}^\mathcal{V},\cdots,P_{d-h+1}^\mathcal{V};\\
&G_g,G_r,G_s
)
\label{equation_problem_definition}
\end{split}
\end{equation}

During a execution, MSTGCN can predict results at network-wide toll stations $\mathcal{V}$ on future $f$ days, which comprehensively considers various spatio-temporal features and extra features. 
The input of our method contains three parts: traffic flow with the dimensions of $V*D$, external feature with the dimension of $V*D*S$, and three graphs of highway network. Here, $V=\lvert\mathcal{V}\rvert$, $D=\lvert\mathcal{D}\rvert$, $S$ is dimensionality of external feature (i.e., $S=2$ according to Section \ref{section4.2}), and $f$ and $h$ represent the time step of prediction and training respectively.

\subsection{Spatio-temporal fusion}\label{section4.4}
The spatio-temporal features would be fused by core stages of MSTGCN as follows. 

{\bfseries Temporal convolution.}
As shown in Figure\ref{structure}, two temporal convolution stages are designed in our method before and after the spatial convolution stage. They have identical structure including a two-dimensional convolution module and a GLU (gated linear units) module, but have different inputs. 
The input of the former is the traffic flow normalized by feature processing stage. The latter deals with the spatio-temporal features produced by spatial convolution stage. The operations here can be expressed by the equation (\ref{equation_temporal_convolution}).

\begin{equation}
\widetilde{X}^{(k)}=\Gamma^{(k)}+\Gamma^{(k)} \odot \Psi^{(k)}
\label{equation_temporal_convolution}
\end{equation}

\begin{equation}
\Gamma^{(k)}=\Theta^{(k)}*Y^{(k)}
\label{equation_t_conv}
\end{equation}

\begin{equation}
\Psi^{(k)}=\sigma(\Theta^{(k)}*Y^{(k)})
\label{equation_glu}
\end{equation}

Hadamard product $\odot$ multiplies  $\Gamma^{(k)}$ and $\Psi^{(k)}$ by elements. Integer $k=\{1, 2\}$ indicates respective stage of temporal convolution: $k=1$ implies the first temporal convolution before the spatial convolution stage; $k=2$ points the second one after the spatial convolution stage. 
Here, $\Gamma^{(k)}$ and $\Psi^{(k)}$ are respectively the results of the convolution module and the GLU module, after the convolution mapping on the input data.

%
Referring to equation (\ref{equation_t_conv}) and (\ref{equation_glu}) in the convolution operation, $Y^{(k)}\in\mathbb{R}^{V*D^{(k)}*S^{(k)}}$ is the normalized values of traffic flow, and $\Theta^{(k)} \in \mathbb{R}^{1*m*C_{in}^{(k)}*C_{out}}$ is convolution kernel. Here, $C_{in}^{(k)}$ and $C_{out}$ are kernel's channel size, and $m$ is a positive integer.  For the input feature $Y^{(k)}$, the temporal features are extracted through $\Theta$ according to nearby $m$ time steps at each toll station, and mapped into $\Gamma^{(k)},\Psi^{(k)} \in \mathbb{R}^{V*(D^{(k)}-m+1)*C_{out} }$. Since the time dimension of the input data is not padding, the time dimension is $D^{(k)}-(m-1)$ here. Eventually, $\widetilde{X}^{(k)} \in \mathbb{R}^{V*D^{(k)-m+1 }*C_{out}}$ is the output of temporal convolution stage.

{\bfseries Spatial convolution.}
In this stage, three graphs defined in Section \ref{section4.3} are employed to fuse their spatio-temporal characteristics for traffic flow at network-wide toll stations. The operation of this stage can be depicted as equation (\ref{equation_graph_fusion}). Over the input $\widetilde{X}^{(1)}$ achieved from the first temporal convolution stage, $\widehat{X}_{fuse} \in \mathbb{R}^{V*D^{(1)-m+ 1}*C_{sout}}$ is the fused spatio-temporal characteristics through graphs of highway network $G_g, G_r$ and $G_s$. 

\begin{equation}
\widehat{X}_{fuse}=\widehat{X}_g+\widehat{X}_r+\widehat{X}_s
\label{equation_graph_fusion}
\end{equation}
\begin{equation}
\widehat{X}=Relu(\hat{D}^{-1/2}\hat{A}\hat{D}^{-1/2}\widetilde{X}^{(1)}\omega_{sc})
\label{equation_graph_gcn}
\end{equation}

For any graph of the threes, graph convolution is adopted to extract spatio-temporal feature $\widehat{X}$ as equation (\ref{equation_graph_gcn}). Here, $\widehat{A}=W+I_V$, where $W$ is $W_g, W_r$ or $W_s$ and $I_V$ represents the unit matrix with dimension $V$. 
The output of graph convolution operation over graphs $G_g, G_r$ or $G_s$ are respectively $\widehat{X}_g$, $\widehat{X}_r$ and $\widehat{X}_s$. In $\hat{D}$, $\hat{D}_{ii}=\sum_j \hat{A}_{ij}$, where $i,j\leq{V}$ and $\omega_{sc}$ is the trainable weight matrix by graph convolution operations. 
The output of spatial convolution stage $\widehat{X}_{fuse}$. $Y^{(2)}=\widehat{X}_{fuse}$ would be the input for the next stage (i.e., the second temporal convolution). 

{\bfseries Full connection.}
In this stage, external factors including calendric type and weather condition defined in Section \ref{section4.2} are employed. Based on fully connected neural network, the spatio-temporal characteristics with external factors are comprehensively considered to predict traffic flow. The procedure is expressed as equation (\ref{equation_fn}). Here, $\omega_{fc}$ and $b$ are trainable parameters: the former is the weight matrix of fully connection stage, and the latter represents the bias.

\begin{equation}
(\hat{y}_{d+1}^V,\hat{y}_{d+2}^V,\cdots,\hat{y}_{d+l}^V)=(P_D^L;\widetilde{X}_{(2)})*\omega_{fc}+b
\label{equation_fn}
\end{equation}

Our method outputs the predictive traffic flow at network-wide toll stations and provides Restful API to acquire those results. Various applications can be facilitated further. Due to the advantage of modularized method, the predictive traffic flow can be easily integrated to enriched current ITS system.

\section{Evaluation}\label{section5}

\subsection{Setting}\label{section5.1}
Has been adopted in a practical ITS mentioned in Section \ref{section3.1}, MSTGCN is evaluated by experiments and a case study in that system. To maintain toll data and aggregative traffic flow data, three virtual machines of our private Cloud form a HBase 1.6.0 cluster, each of which owns 4 cores CPU, 22 GB
RAM and 700 GB storage. Another machine (2 cores Intel Xeon W-2125 CPU, 8 GB RAM, 200 GB storage, and GPU NVIDIA GeForce RTX 2080 Ti) installing CentOS 6.6 x86\_64 operating system is used to build MySQL 5.6.17 as the relational database for both business profiles (station, section and highway line) and external data (i.e., calendric data and weather data). Our method is developed on that machine by Python 3.6, PyTorch 1.9 and torch-geometric 1.7.2. 

Daily traffic flow data at all the 269 toll stations of highway network are employed since May 2017 to September 2017. The data on the latest 15 days of September 2017 is used as test set, and the rest of data is regarded as training set. The time steps of training and prediction are respectively set as $h=15$, $f=1$. In temporal convolution stage, the parameters of the convolution kernel are set to $C_{out}=64$ and $m=3$. In spatial convolution stage, the parameter of graph convolution module is set to $C_{sout}=16$. 

{\bfseries Evaluation metric $\&$ baselines:} In order to evaluate the predictive effect quantitatively , three commonly used metrics for regression problems are adopted here, including root mean square error (RMSE), mean absolute percentage error (MAPE) and mean absolute error (MAE). Their equations are (\ref{equation_rmse} - \ref{equation_mae}) respectively. Here, on a certain day, $V$ represents the size of toll stations $\mathcal{V}$ in highway network; at at toll station $v\in\mathcal{V}$,  $\hat{y}^{v}$ is the predictive traffic flow, and $y^{v}$ is the ground truth.

\begin{equation}
RMSE=\sqrt{\frac{1}{V}\sum^V_{v=1} (\hat{y}^{v}-y^{v})^2}
\label{equation_rmse}
\end{equation}
\begin{equation}
MAPE=\frac{100\%}{V}\sum^V_{v=1}|\frac{\hat{y}^{v}-y^{v}}{y^{v}}|
\label{equation_mape}
\end{equation}
\begin{equation}
MAE=\frac{1}{V}\sum^V_{v=1}|\hat{y}^{v}-y^{v}|
\label{equation_mae}
\end{equation}

In addition, we compared our work with several baselines used in highway domain: Hybrid Graph Convolutional Network (HGCN)~\cite{ref_article34}, Spatio-Temporal Graph Convolutional Networks (STGCN)~\cite{ref_article18}, and traditional machine learning models GBRT~\cite{ref_article7} and KNN~\cite{ref_article6}.

\subsection{Experiment}\label{section5.2}
Two experiments are designed for quantitative evaluation. In the first experiment, our method's predictive accuracy is compared with baselines at network-wide toll stations. In the second one, spatio-temporal factors and their affect in our method are discussed.

{\bfseries Experiment 1: predictive accuracy comparison with baselines.} In the system above, our method MSTGCN is to predict daily traffic flow at network-wide toll stations of Henan province, as the settings in Section \ref{section5.1}. Its predictive accuracy is compared with the baselines. Here, the hyper parameters of HGCN and STGCN can refer to~\cite{ref_article34}~\cite{ref_article18} respectively; for KNN~\cite{ref_article6}, neighbour size parameter $k=5$; in GBRT~\cite{ref_article7}, tree size is $M=3000$ and maximal tree depth $d=3$. Their average predictive accuracy at network-wide toll stations are showed in table~\ref{tab_1}.

\begin{table}[H]
\begin{center}
\caption{Average performance at network-wide toll stations.}\label{tab_1}
\setlength{\tabcolsep}{0.5\tabcolsep}{
\begin{tabular}{l||lll}
\hline
\hline
Model &  MAE & MAPE(\%) &RMSE\\
\hline
\hline
KNN&303.686&17.037&573.912\\
GBRT&246.359&11.580&459.467\\
HGCN&403.861&23.185&682.393\\
STGCN&239.849&8.589&468.486\\
MSTGCN&\textbf{197.421}&\textbf{6.936}&\textbf{405.931}\\
\hline
\end{tabular}}

\end{center}
\end{table}

From table~\ref{tab_1}, we find that MSTGCN has achieved the best performance in all the three metrics. Taking MAE metric for example, MSTGCN has reduced by at least 17\% than others. Such effects comes from three aspects. First, feature processing stage of our method reduces the deviation by reasonable data normalization. It alleviates the predictive difficulties on the massive data in imbalanced distribution. Second, various semantics from three graphs are employed, and more comprehensive spatio-temporal characteristics can be fused. The sub-optimal baseline STGCN also adopts graph convolution network, but it builds graph structure only from physical highway topology. Such static and partial spatial feature makes its performance worse than ours. Third, external factors imported in full connection stage, is significant to reflect key periodic information. That is why MSTGCN obviously has advantageous performance than baseline HGCN, KNN and GBRT. 

\begin{figure*}[htbp!]
\begin{center}
  \includegraphics[width=0.5\textwidth, height=0.25\textwidth]{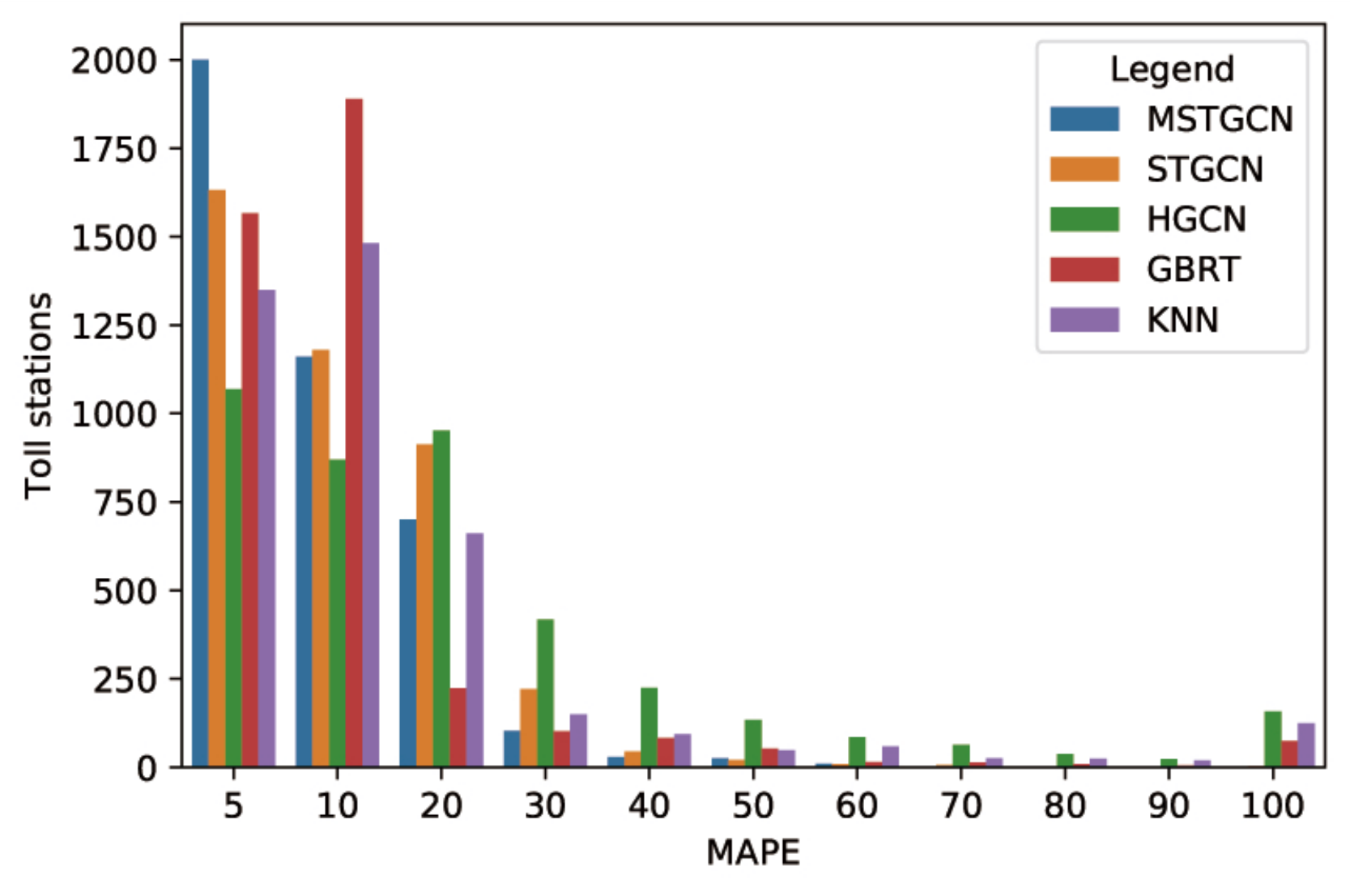}
  \caption{MAPE distribution at network-wide toll stations}
  \label{mape_distribution}
\end{center}
\end{figure*}

Such effects can also be analyzed in another view. The average of metric MAPE over the days of test data is calculated at any toll station, and the MAPE distribution at network-wide toll stations can be found in Figure \ref{mape_distribution}. The toll station, whose average is not larger than the value of x-coordinate, would be counted as the accumulative value of y-coordinate. Two interesting evidences are found. On the one hand, through any of the five models, the predictive results own relatively low errors. Most MAPEs at toll stations are not larger than 20\%. On the other hand, MSTGCN and STGCN perform better than others due to visible positive-skew (right-skew) distribution. In fact, MSTGCN is the best from the facts: its first bucket (i.e., with the lowest error) has largest count, and the first two buckets (i.e., with MAPE smaller than 5\%) contains almost 50\% among all the toll stations. 
HGCN is the worst in general because its graph of highway network only depicts physical semantics. GBRT performs a little better than KNN, but both have several toll stations whose MAPE is larger than 40\%. In fact, just the "vital few" toll stations would own such bad predictive results. 

In order to show convincing merits of our method with insight, several variants of MSTGCN and the sub-optimal STGCN are configured for detailed ablation experiment as follows.

{\bfseries Experiment 2: Spatio-temporal factors analyses of MSTGCN.}  As variants of our method, $\rm{MSTGCN_{nonT}}$ removes feature processing stage from the original MSTGCN; $\rm{MSTGCN_{gs}}$ only considers geographic graph and elastic graph in spatial convolution stage; $\rm{MSTGCN_{rs}}$ only employ influential graph and elastic graph in spatial convolution stage; $\rm{MSTGCN_{nonE}}$ removes external factors in full connection stage. Through those variants, original MSTGCN and STGCN, their average predictive accuracy at network-wide toll stations are evaluated from two perspectives. One is to show temporal effects on typical weekday and weekend. The weekday Sep.21 2017 and the weekend Sep.24 2017 are chosen here, and results are illustrated in Table \ref{table_days}. The other is to present spatial effects at typical toll stations with different volume of traffic. The toll stations \textit{Zhengzhouhouzhai} and \textit{Anyang} are selected here: the former has large traffic (more than 10000), whose daily traffic flow is about 11000; the latter has small traffic (less than 10000), whose traffic flow is about 4000. The results are presented in Table \ref{table_stations}.

\begin{table*}
\caption{Performance on typical days}
\centering
\label{table_days}
\setlength{\tabcolsep}{0.5\tabcolsep}{
\begin{tabular}{l||lll|lll}
\hline
\hline
& RMSE & MAE& MAPE(\%) & RMSE&MAE&MAPE(\%)\\

&\multicolumn{3}{|c|}{20170921(weekday)}&\multicolumn{3}{|c}{20170924(weekend)}\\
\hline
\hline
STGCN&293.514&182.146&6.224&378.099&252.419&11.130\\
$\rm{MSTGCN_{nonT}}$&256.607&165.810&6.560&320.085&220.686&10.394\\
$\rm{MSTGCN_{gs}}$&291.053&153.958&5.387&319.300&191.421&8.636\\
$\rm{MSTGCN_{rs}}$&232.945&141.100&5.745&277.109&188.862&8.991\\
$\rm{MSTGCN_{nonE}}$&274.817&153.612&5.312&385.730&196.377&7.577\\
MSTGCN&\textbf{185.714}&\textbf{116.437}&\textbf{4.584}&\textbf{269.210}&\textbf{165.348}&\textbf{7.350}\\
\hline
\end{tabular}}
\end{table*}
From temporal effects in Table \ref{table_days}, three key facts can be found. 
First, the feature processing is significant for traffic flow prediction on both weekday and weekend. It is clearly reflected by the worse performance than $\rm{MSTGCN_{nonT}}$ among $\rm{MSTGCN}$ and variants. Data normalization of our method can alleviate predictive errors at certain toll stations in a imbalanced data distribution. 
Second, calendric type as periodic external factor benefits the performance. Almost in all the method/models, the performance on weekday is better than that of weekend. It comes from the fact that traffic flow appears more regular patterns on weekday than on weekend. Not considering calendric external factor, STGCN obviously performs worse than others. It is the same reason that $\rm{MSTGCN_{nonE}}$ shows poor effect on weekend than other variants of MSTGCN. 
Third, the more comprehensive spatial semantics are employed, and the better prediction would be. It can be found on both weekday and weekend, MSTGCN shows better results than $\rm{MSTGCN_{gs}}$ and $\rm{MSTGCN_{rs}}$, which learns more from three different graph of highway network.  
In brief, by fusing external factors and spatio-temporal feature, MSTGCN proves the best predictive accuracy on both weekday and weekend. Compared with STGCN, the metric RMSE of MSTGCN has reduced by nearly 30\%.

The spatial effects in Table \ref{table_stations} show that traffic flow appears different patterns at specific toll stations. On the one hand, the result implies that spatial factors matter much for traffic flow prediction. $\rm{MSTGCN_{rs}}$ owns highest predictive accuracy at the toll station with large traffic. It implies the influential graph is more dominant at such toll stations than other graphs of highway network. Although slightly better than MSTGCN, predictive accuracy of $\rm{MSTGCN_{rs}}$ is still in the same level with MSTGCN. At the toll station with small traffic, MSTGCN performs almost the same as $\rm{MSTGCN_{rs}}$. On the other hand, multi-graph convolution proves its benefit to capture the spatial feature of highway network. $\rm{MSTGCN}$ and its variants perform much higher than STGCN at either toll station, where STGCN only employs static geographic graph of highway network. 
The predictive effects on days of two weeks can be also reflected in Figure \ref{fig_flow_status}, our $\rm{MSTGCN}$ and $\rm{MSTGCN_{rs}}$ fits the ground truth better especially at the large traffic toll station. 
In brief, MSTGCN performs high enough predictive accuracy by captured spatial feature from multiple perspectives.
\begin{table*}
\caption{Performance at typical locations}
\centering
\label{table_stations}
\setlength{\tabcolsep}{0.5\tabcolsep}{
\begin{tabular}{l||lll|lll}
\hline
\hline
& RMSE & MAE& MAPE(\%) & RMSE&MAE&MAPE(\%)\\

&\multicolumn{3}{|c|}{Zhengzhouhouzhai}&\multicolumn{3}{|c}{Anyang}\\
&\multicolumn{3}{|c|}{(large traffic toll station)}&\multicolumn{3}{|c}{(small traffic toll station)}\\
\hline
\hline
STGCN&1357.367&1124.394&7.570&429.760&350.383&5.482\\
$\rm{MSTGCN_{gs}}$&1085.348&890.934&5.704&452.559&367.827&5.553\\
$\rm{MSTGCN_{rs}}$&\textbf{815.510}&\textbf{653.416}&\textbf{4.272}&331.402&\textbf{250.585}&\textbf{3.793}\\
$\rm{MSTGCN_{nonE}}$&991.632&822.948&5.225&387.564&291.165&4.365\\
MSTGCN&823.322&677.603&4.474&\textbf{324.154}&254.071&3.877\\
\hline
\end{tabular}}
\end{table*}

{\fontfamily{pcr}\selectfont
\begin{figure*}[htbp]
\begin{center}
\includegraphics[width=\textwidth, height=0.25\textwidth]{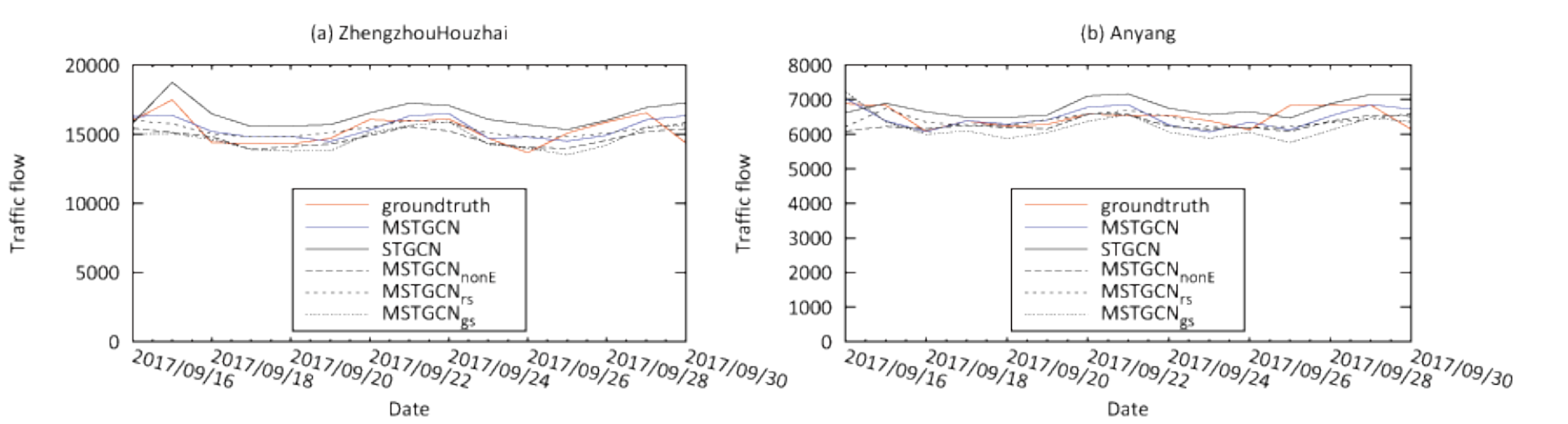}
\caption{Predictive effects on continuous days at typical toll stations} \label{fig_flow_status}
\end{center}
\end{figure*}
}

\subsection{Case study}\label{section5.3}
In the system mentioned in Section\ref{section3.1}, currently 269 toll stations of Henan highway have been managed. Our method executes at 12:00 a.m. everyday to predict network-wide traffic flow for a coming day. All the results would be written to HBase storage. In this sub-section, two applications in that system are explained as case studies of MSTGCN. One is highway hot-spot detection among toll stations, and the other is travel trend analysis on specific holidays.

\begin{figure*}[htbp]
\begin{center}
\includegraphics[width=0.7\textwidth, height=0.4\textwidth]{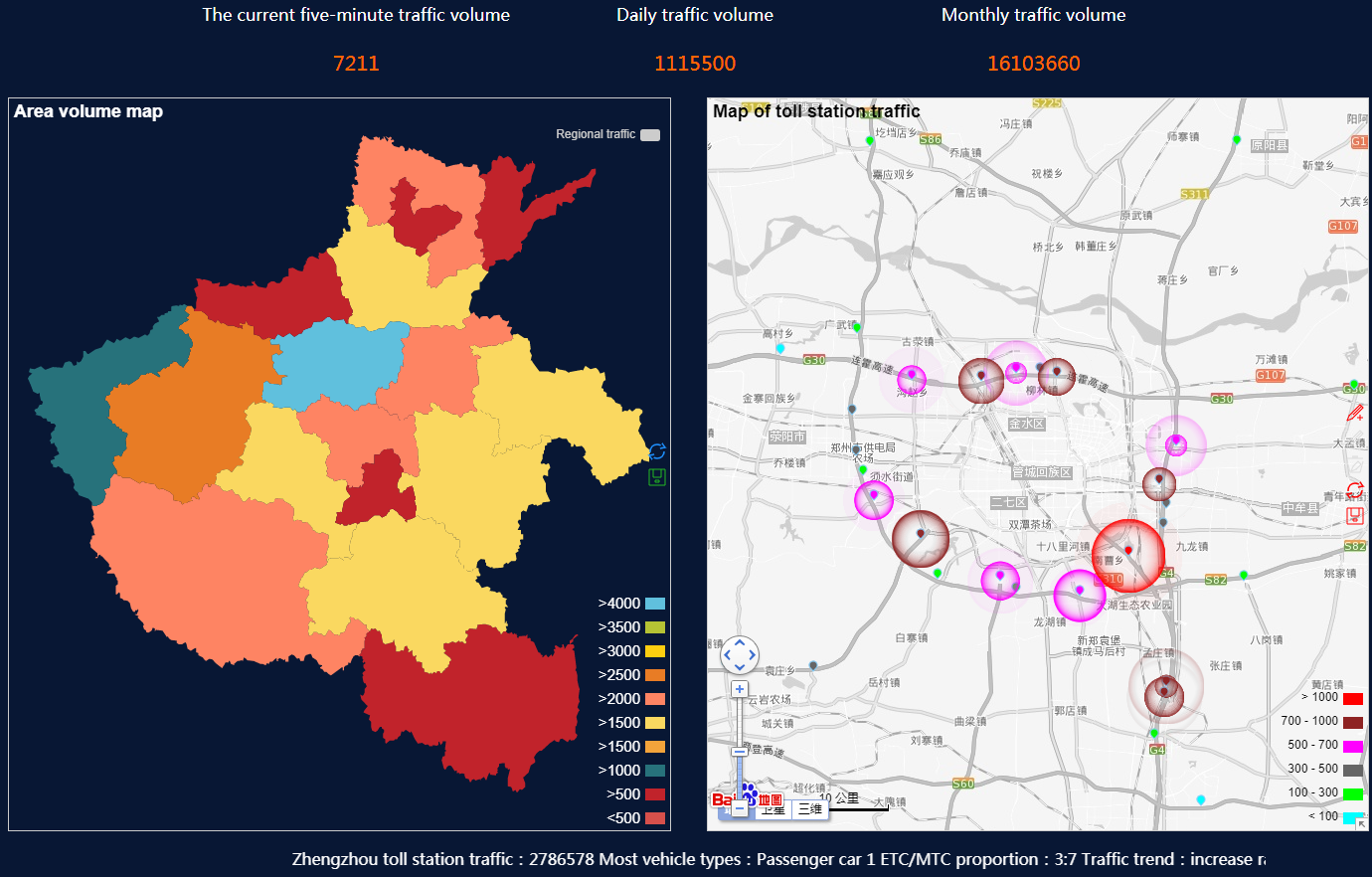}
\caption{Potential traffic hot-spots among toll stations} \label{fig_hotspots}
\end{center}
\end{figure*}

{\bfseries Case 1}. In highway network, potential traffic hotspots can be found by evaluating future traffic flows at network-wide toll stations~\cite{ref_article7}. Based on our method, daily hot-spot detection application in the system is showed as Figure \ref{fig_hotspots}. The left is an administrative map of \textit{Henan} province, where heat degree of cities are presented in regional colours. The right is a map of provincial city \textit{Zhengzhou} with surrounding toll stations, where heat degree of stations is noted in colourful bubbles. In either map, the potential hot-spots of stations on a coming day are intuitive by data visualization. 

Traffic hot-spots among toll stations implies heavy traffic, which is closely concerned in business. By heat degrees in both maps, the ordered traffic flows at toll stations are visualized. Heat degree of a prefecture-level city in the left map implies a sum of traffic flows at the toll stations located in that city. According to quantities of that sum, such heats appear in different colours. We found the provincial city \textit{Zhengzhou}, which is located in central north of the province, has the largest amount of traffic flow in blue colour. After clicking a city in left map, we find stations of that city in right map. City \textit{Zhengzhou} is shown in the right of Figure \ref{fig_hotspots}, where ten more toll stations surround its peripheral. The potential hot-spots of toll stations are displayed as colourful bubbles in different diameters. The darker the colour is, the bigger the bubble appears, and the larger the predicted traffic flow at a toll station would be. Here in Figure \ref{fig_hotspots}, we can directly find trends that stations in east and south of city \textit{Zhengzhou} would be busy, because seven stations there are potential hot-spots. Our method is employed to predict traffic flow at network-wide toll stations. In fact, the vital few stations in long-tail distribution are always hot-spots, and our method significantly keeps their predictive accuracy and catches the domain meaning for business technicians with interactive visual maps. 

\begin{figure*}[htbp]
\begin{center}
\includegraphics[width=0.7\textwidth, height=0.4\textwidth]{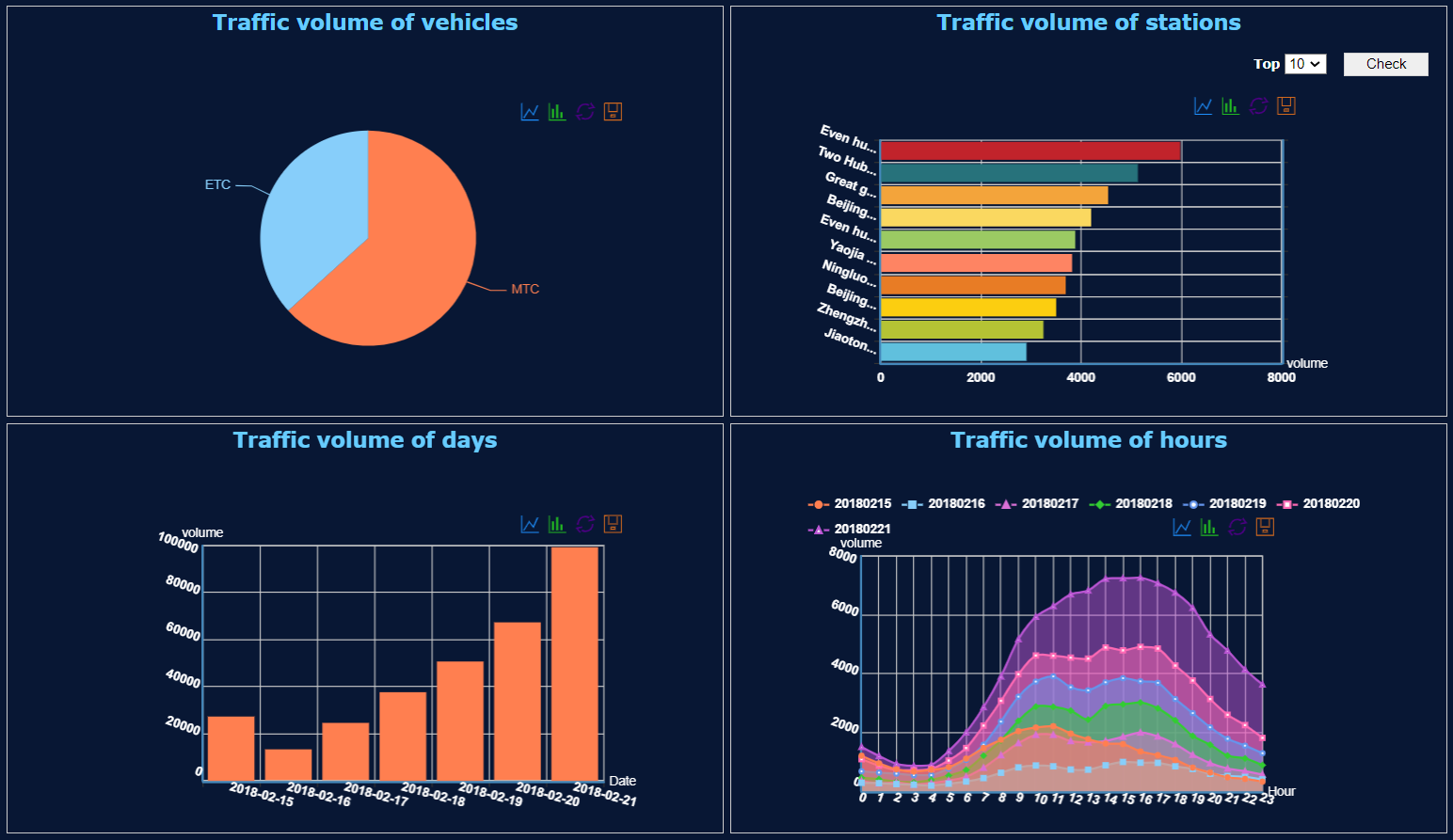}
\caption{Potential traffic hot-spots among toll stations} \label{fig_trends}
\end{center}
\end{figure*}

{\bfseries Case 2}. As the discussion in Section \ref{section4.2}, calendric type like holidays always implies respective traffic pattern in highway. On some of holidays, toll-free policy would be carried out by Chinese Ministry of Transport, and possible burst of private travel makes much highway stress on those days. Based on our method, a trends-subject application for Spring Festival 2018 in our system is shown as Figure \ref{fig_trends}. On this 7-day national holiday, the predicted traffic flows are represented in four perspectives: vehicular type proportion, toll station ranking, daily comparison, and hourly comparison on each date. 

Traffic trends are represented in four perspectives in Figure \ref{fig_trends}, which are supported by network-wide traffic flow. In the first perspective located left-top, predicted daily traffic flows are summarized from all the stations on those seven days, and then divided into two types by driver identity (i.e., either MTC or ETC). Such drivers’ entity characteristics are reflected by a proportion of identities. In the second perspective located right-top, toll stations are ordered by their summary of predicted traffic flows on those seven days. Spatial characteristics are ranked by a bar chart, and top-10 stations are presented in descending order here. In the third perspective located left-bottom, respective dates with the summary of network-wide traffic flows are compared. Temporal characteristics are reflected by a histogram. In the fourth perspective located right-bottom, hours are compared by short-term prediction on each of the seven days. Referred from our previous work ~\cite{ref_article6}, fine-granularity temporal characteristics on respective day can be clearly found: traffic peaks on the last two days of that holiday are prominently high than others, because return flows back to drivers’ locale would bust intensively when a holiday is close to the end. Accordingly, our method proves its extensive feasibility and availability in practice.

\begin{figure*}[ht]
\begin{center}
\includegraphics[width=0.8\textwidth, height=0.3\textwidth]{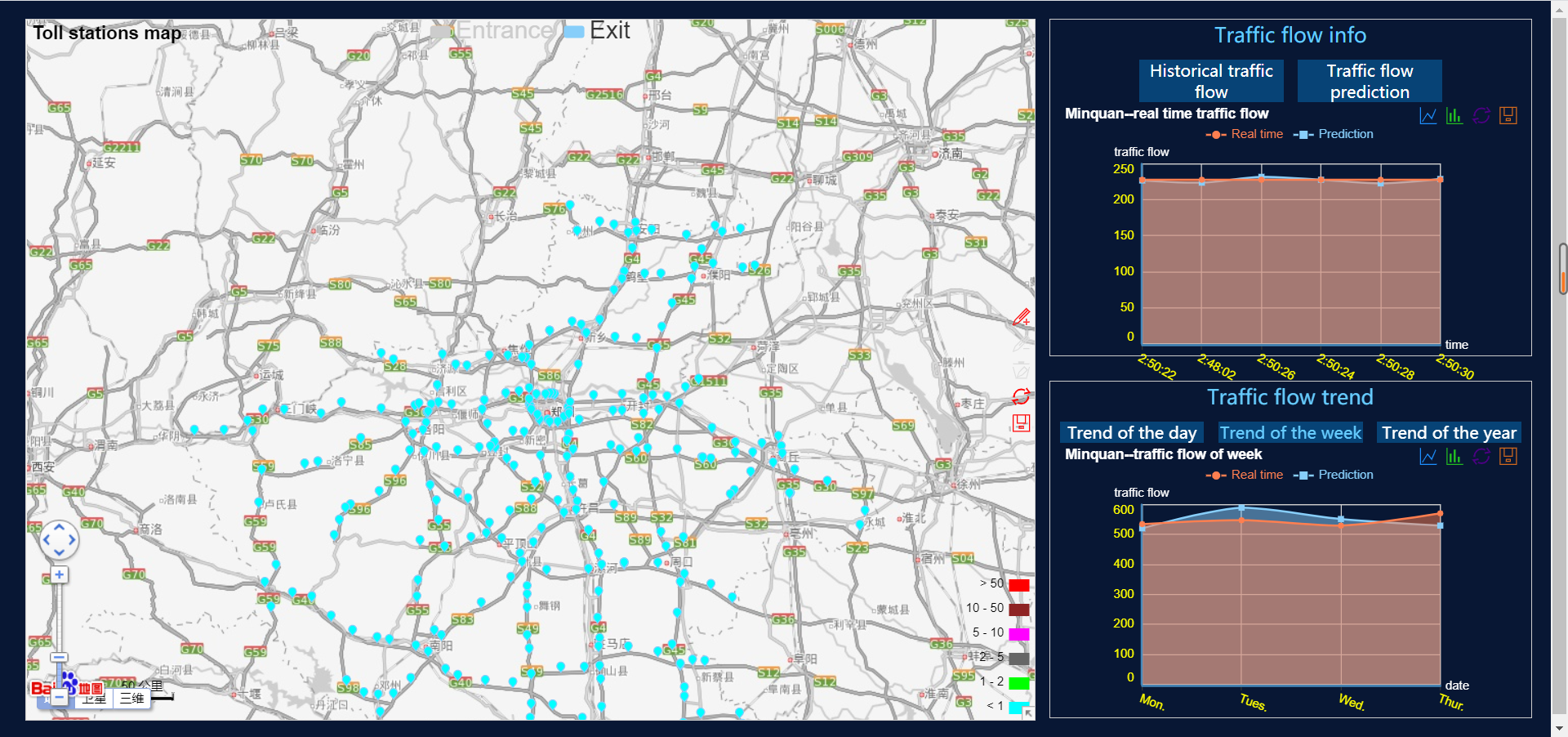}
\caption{Toll station traffic flow monitoring and prediction} \label{fig_prediction}
\end{center}
\end{figure*}

{\bfseries Case 3}. A dashboard for traffic flow prediction of the system is presented in Figure \ref{fig_prediction}. A provincial map of Henan is in the left part with all toll stations dotted. It would be interactive with the right part for predicted traffic flow. When a station is selected in the map, its traffic flow would be presented in two right charts. The upper chart is the real-time traffic flow with its prediction; the bottom one shows the weekly perspective (composed by daily predictive values). Figure \ref{fig_prediction} here shows the traffic flow at toll station $Minquan$. In both right charts, the yellow line represents ground truth of traffic flow, and the blue line is predictive results generated by our method. 

We can show insight about MSTGCN in this case. Among three highway network graphs of our method, geographic graph is built just as the physical map topology; influential graph is constructed through historical statistics on traffic flows at toll stations; elastic graph is dynamically self-learned to find inherent relationship. Take the toll station $Minquan$ in the map here as an example. In geographic graph, toll stations $Lankao, Erlangmiao$, and $Ningling$ can be reached directly in map; in influential graph, top three toll stations are $Xunchangbei, Nansanhuan, Shangquexi$ by historical statics on historical traffic flows; in elastic graph, top three toll stations are $Yanshi, Yongcheng, Xiping$ by self-learning. Therefore, the graphs from different perspectives reflect comprehensive spatial semantics in highway network. With those graphs, accurate predictive results can be achieved through spatio-temporal fusion in this system. 

\section{Conclusion}\label{section6}
In this paper, a method MSTGCN through multi-graph spatio-temporal fusion is proposed for daily traffic flow prediction. The spatial characteristics of traffic flow are trained in three graphs of highway network. After the feature processing for data in imbalanced distribution, the spatio-temporal feature and external factors are comprehensively fused to obtain more accurate predictive results. 
By extensive experiments and case studies in one Chinese provincial highway, our method shows at least 13\% reduction by metric RMSE than baselines, proves convincing benefits in practice.

%
%
%

\bibliographystyle{unsrt}

\bibliography{my_paper}




\end{document}